\documentclass{article}
\PassOptionsToPackage{numbers, compress}{natbib}
\usepackage[preprint]{neurips_2025}

\usepackage{url}
\usepackage{hyperref}
\usepackage{booktabs}
\usepackage{amsfonts}
\usepackage{nicefrac}
\usepackage{microtype}
\usepackage{amsmath}
\usepackage{xcolor}
\usepackage{algorithm}
\usepackage{algpseudocode}
\usepackage{amssymb}
\usepackage{graphicx}
\usepackage{float}
\usepackage{tikz}
\usepackage{pgfplots}
\pgfplotsset{compat=newest}
\usepackage{tcolorbox}
\usepackage{listings}
\usepackage{listingsutf8}
\usepackage{caption}
\usepackage{subcaption}
\usepackage{contour}
\usepackage{multirow}
\newcommand{\Q}{\btt{Q}}
\newcommand{\A}{\btt{A}}
\newcommand{\btt}[1]{{\fontseries{sb}\selectfont\scalebox{1.00}[1.00]{#1}}}
\title{PASTA: A Paraphrasing And Self-Training Approach for Knowledge Updating in LLMs}

\author{
Takayuki Yamamoto \\
Waseda University \\
\texttt{takayukiyamamoto@ruri.waseda.jp}
\and
Daisuke Kawahara \\
Waseda University \\
\texttt{dkw@waseda.jp}
}

\makeatletter
\renewcommand{\@noticestring}{}
\makeatother

\begin{document}

\maketitle
\begin{abstract}
Knowledge updating in pre-trained Large Language Models (LLMs) remains an important challenge. While continual training provides a potential avenue for knowledge updating, it continues to present substantial technical difficulties. Furthermore, LLMs often struggle with accurately answering questions about specific factual information, such as news articles—a capability limitation widely recognized in the research community. This paper proposes PASTA, a simple yet powerful framework for integrating detailed factual information from news articles as new knowledge into LLMs, with the primary goal of building specialized models that accurately answer questions about this knowledge. Our framework combines data augmentation, question-answering generation, and a novel self-learning DPO process that simultaneously enables knowledge overwriting and hallucination suppression. We provide insights into effective knowledge updating through systematic analysis of learning parameters and data configurations. In our experimental evaluation with web articles published after the base model's knowledge cutoff, PASTA achieved remarkable improvement from 0.02 to 0.82 accuracy while maintaining general language capabilities, demonstrating its effectiveness for creating domain-specialized LLMs.
\end{abstract}

\section{Introduction}
Updating pre-trained Large Language Models (LLMs) with new knowledge and training them to respond accurately based on those facts is recognized as a significant challenge \cite{wei2024measuring,kuroki2024agent}.
While approaches such as Retrieval-Augmented Generation (RAG) \cite{lewis2020retrieval} have been researched as a method to provide new knowledge to language models, this paper focuses on methods for direct internal knowledge updating, aiming to build LLMs capable of correctly answering questions about newly integrated information. Through our proposed approach, we demonstrate that an LLM which initially showed only 0.02 accuracy when answering questions about unknown knowledge can achieve up to \btt{0.82 accuracy} after knowledge updating.
Previous research has suggested that diverse training data is closely related to accurate knowledge retrieval by probing internal representations of LLMs \cite{allen2023physics}.
However, this research has only examined cases where relatively small-scale models \cite{radford2019language} were trained from scratch, without sufficiently investigating additional training with large language models as initial parameters.
On the other hand, methods have been proposed to build specialized LLMs for specific tasks (e.g., programming) through additional training using large corpora \cite{roziere2023code}.
These approaches supplement relatively broad and static knowledge (programming languages, algorithms, etc.), but challenges remain in developing methods for LLMs to learn and accurately answer questions about specific, detailed factual information such as news articles.

This paper proposes PASTA, a simple yet powerful framework for knowledge updating of small-scale detailed factual information, such as news articles, on a pre-trained large language model through continual learning. While each individual component builds upon established techniques, their systematic combination creates a novel approach that significantly outperforms baseline methods. The simplicity of PASTA's design makes it both practical to implement and theoretically interpretable, while demonstrating substantial performance gains in knowledge updating tasks.

Drawing inspiration from recent self-learning approaches \cite{li2023self,yuan2024self}, we combine the following four steps:
(1) data augmentation \cite{allen2023physics,brown2024large},
(2) question-answering (QA) data generation \cite{li2023self},
(3) self-preference optimization \cite{yuan2024self}, and
(4) iterative learning \cite{li2023self}.
We present a reconstructed additional learning flow that integrates these components and experimentally confirm that new knowledge can be gradually incorporated.

\section{Related Work}
\label{sec:related_work}
\subsection{Preference Optimization}
\label{Preference Optimization}
Representative methods for optimizing LLMs to align with human preferences include
Reinforcement Learning from Human Feedback (RLHF) \cite{ouyang2022training,christiano2017deep} and
Direct Preference Optimization (DPO) \cite{NEURIPS2023_a85b405e}.
Both methods share the common approach of presenting two output candidates from the model and labeling them as ``chosen'' or ``rejected''. However,
RLHF involves training a reward model followed by reinforcement learning, while DPO directly optimizes parameters without explicitly training a reward model.
\subsection{Self-Learning}
\label{subsec:self_learning}
Large-scale human annotation is costly, and when it includes sensitive user information, privacy concerns also arise. Therefore, self-learning approaches, where the model generates and evaluates its own data for iterative learning, have gained attention.
For instance, \citet{li2023self} reported improved model performance by automatically generating diverse QA data for retraining.
\citet{yuan2024self} proposed "self-DPO learning" within the DPO framework \cite{NEURIPS2023_a85b405e}, where models learn by distinguishing between high-scoring and low-scoring responses generated by the model itself.
\citet{wang2024self} introduced mechanisms for LLMs to autonomously conduct pairwise comparisons.
\citet{lee2024rlaif} explored using LLMs as partial substitutes for human annotation.
On the other hand, \citet{madaan2024self} and \citet{wang2024languagemodelscollapsetrained} have shown that performance may degrade under certain conditions when models are repeatedly retrained solely on self-generated data, highlighting the importance of appropriate data diversity and validation.

\subsection{Data Augmentation}
\label{subsec:data_augmentation}
While large volumes of data are effective for improving LLM performance, introducing diversity into the data is also beneficial beyond simple scaling.
\citet{allen2023physics} revealed that generating diverse paraphrases for training can better integrate new knowledge, and that using relatively high LoRA \cite{hu2021lora} ranks ($r=128$) can also improve performance.
\citet{brown2024large} also reported that creating multiple variations for a single piece of data and training repeatedly significantly improves reasoning accuracy.

\section{Proposed Method}
\begin{figure}[t]
\centering
\includegraphics[width=0.9\linewidth]{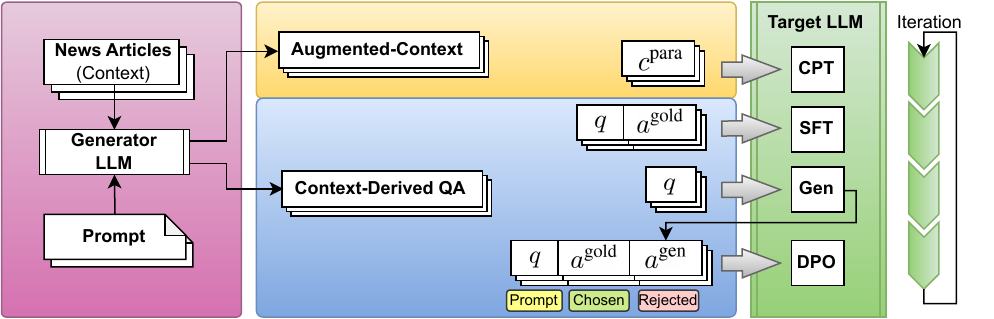}
\caption{PASTA framework architecture. CPT: Continual Pre-Training, SFT: Supervised Fine-Tuning, DPO: PASTA's specialized Direct Preference Optimization with Target LLM inference (Gen).}
\label{fig}
\end{figure}
In this paper, we propose a Paraphrasing And Self-Training Approach (PASTA) as a framework for effectively integrating new knowledge into pre-trained large language models. PASTA consists of four key components: (1) data augmentation through paraphrasing, (2) question-answering data generation, (3) self-preference optimization, and (4) iterative learning based on fundamental LLM development principles. As shown in Figure~\ref{fig}, it forms an integrated framework that iteratively enhances an LLM's ability to accurately answer questions about new knowledge that has been integrated through knowledge updating.

\subsection{Design Motivations}
\label{sec}
\paragraph{Diversifying Training Data through Paraphrasing}
LLMs have a structurally redundant architecture where a single meaning is not represented by specific parameters alone, but rather by multiple parameter sets throughout the model \cite{han2015learning,frankle2018lottery}.
Therefore, when integrating new knowledge, simply repeating the same text may concentrate parameter updates in specific areas, failing to update all necessary parameters in the redundant structure, potentially making knowledge updating ineffective.
We hypothesize that learning with diverse paraphrases of the target text—maintaining the same meaning but varying the expression—will update all necessary parameters across the redundant structure, making knowledge updating in LLMs more effective.

\paragraph{Scaling Question-Answering}
When updating knowledge in LLMs, there is difficulty not only in directly updating the knowledge itself but also in propagating updates to related knowledge. For example, when a music artist releases a new album, the total number of albums they have released also increases \cite{hase2024fundamental}. To address these challenges, we need to learn answers to diverse questions about the new knowledge and related topics from multiple angles. We hypothesize that scaling such question-answering will improve the accuracy of knowledge updating.

\paragraph{Hallucination Suppression Using DPO}
LLMs sometimes produce hallucinations, presenting non-existent information as if it were factual \cite{farquhar2024detecting}.
Particularly for detailed factual information such as news articles, where the correct answer to the same question may change over time (e.g., a president's name differs depending on the time period), there is a risk of generating incorrect answers using outdated knowledge from before knowledge updating.
Therefore, we hypothesize that by using DPO to identify outputs based on the LLM's inherent knowledge (old information) as "rejected" responses and newly added knowledge as "chosen" responses, we can suppress inaccurate outputs.

\paragraph{Pre-training + SFT + DPO Framework}
A widely adopted three-stage learning pipeline for LLM development includes (1) pre-training, (2) SFT, and (3) final adjustment to human preferences (using methods like RLHF or DPO) \cite{ouyang2022training}.
In this work, we aim to achieve effective knowledge updating and build an LLM capable of accurately answering questions about that content by integrating the above hypotheses into a system based on this three-stage learning pipeline.

\subsection{Formulation of the PASTA Framework}

To formalize the PASTA framework, we first introduce basic notation. We denote the set of new knowledge, such as news articles, as $\mathcal{C} = \{c_1, c_2, \ldots, c_K\}$, where $K$ is the total number of contexts (articles). For each article $c_k$ (where $k \in \{1,2,\ldots,K\}$ is the article index), we provide it to the \btt{Generator LLM} $G$ along with an instruction prompt (see Appendix~\ref{appendix:Augmented-Context_prompt}) to generate $P$ paraphrase variations. This set of paraphrases is denoted as $\mathcal{C}_k^{\text{para},i} = \{c_{k,1}^{\text{para},i}, c_{k,2}^{\text{para},i}, \ldots, c_{k,P}^{\text{para},i}\}$ and referred to as \btt{Augmented-Context} in this paper. The subscript $i$ indicates the iteration number, and the superscript ``para'' indicates that the context has been paraphrased. The complete set of paraphrases is defined as $\mathcal{C}^{\text{para},i} = \{c_{k,p}^{\text{para},i} \mid k \in \{1,\ldots,K\}, p \in \{1,\ldots,P\}\}$ and is used for Continual Pre-Training (CPT).
Additionally, we generate $Q$ question-answer pairs from each article, represented as $\mathcal{Q}_k^i = \{(q_{k,j}^i, a_{k,j}^{i,\text{gold}})\}_{j=1}^{Q}$. Here, $q_{k,j}^i$ is the question and $a_{k,j}^{i,\text{gold}}$ is the correct answer. These question-answer pairs are generated by providing the article $c_k$ and an instruction prompt (see Appendix~\ref{appendix:prompt_qa_generation}) to the \btt{Generator LLM} $G$, so $a_{k,j}^{i,\text{gold}}$ is labeled as ``gold'' to indicate a high-accuracy answer based on the article content. Generating numerous question-answer pairs also leads to an increase in similar QA pairs, which is expected to have an effect equivalent to diverse paraphrasing. The complete set of question-answer pairs generated this way, $\mathcal{Q}^i = \{(q_{k,j}^i, a_{k,j}^{i,\text{gold}}) \mid k \in \{1,\ldots,K\}, j \in \{1,\ldots,Q\}\}$, is defined as \btt{Context-Derived QA} in this paper. This question-answer dataset is used for both Supervised Fine-Tuning (SFT) and Direct Preference Optimization (DPO) learning stages, as well as for evaluation at each learning stage by sampling. Furthermore, new \btt{Context-Derived QA} is generated at the end of each iteration to evaluate generalization performance and to be used as training data for the next iteration.
We denote the \btt{Target LLM} to be updated as $M_\theta$, where $\theta$ represents the model parameters. The model parameters at the end of iteration $i$ are denoted as $\theta^i$. The loss functions used in each learning stage are $\mathcal{L}_{\text{CPT}}$, $\mathcal{L}_{\text{SFT}}$, and $\mathcal{L}_{\text{DPO}}$. The number of evaluation samples is $S$, and the total number of iterations is $T$.

\begin{algorithm}[t]
\small
\caption{PASTA: Main Learning and Evaluation Process}
\label{alg:main_process}
\setlength{\baselineskip}{0.85\baselineskip}
\begin{algorithmic}[1]
\linespread{0.9}
\State \footnotesize // Data generation phase prepares $\mathcal{C}^{\text{para},i}$ (\btt{Augmented-Context}) and $\mathcal{Q}^i$ (\btt{Context-Derived QA})
\State \footnotesize $\mathcal{Q}^i_{\text{iter\_sample}} \gets \text{SampleQA}(\mathcal{Q}^i, S)$ \Comment{\footnotesize Samples for evaluation}
\For{\footnotesize $i = 1$ to $T$} // Learning process for each iteration
    \State \footnotesize $\Delta\theta_{\text{CPT}}^i \gets \underset{\Delta\theta}{\arg\min} \sum_{c^{\text{para}} \in \mathcal{C}^{\text{para},i}} \mathcal{L}_{\text{CPT}}(M_{\theta^{i-1}+\Delta\theta}, c^{\text{para}})$ \Comment{\footnotesize LoRA training}
    \State \footnotesize $\theta_{\text{CPT}}^i \gets \theta^{i-1} + \Delta\theta_{\text{CPT}}^i$; $\text{acc}_{\text{CPT}}^i \gets \text{Evaluate}(M_{\theta_{\text{CPT}}^i}, \mathcal{Q}^i_{\text{iter\_sample}})$

    \State \footnotesize $\Delta\theta_{\text{SFT}}^i \gets \underset{\Delta\theta}{\arg\min} \sum_{(q, a^{\text{gold}}) \in \mathcal{Q}^i} \mathcal{L}_{\text{SFT}}(M_{\theta_{\text{CPT}}^i+\Delta\theta}, q, a^{\text{gold}})$ \Comment{\footnotesize LoRA training}
    \State \footnotesize $\theta_{\text{SFT}}^i \gets \theta_{\text{CPT}}^i + \Delta\theta_{\text{SFT}}^i$; $\text{acc}_{\text{SFT}}^i \gets \text{Evaluate}(M_{\theta_{\text{SFT}}^i}, \mathcal{Q}^i_{\text{iter\_sample}})$

    \State \footnotesize $\mathcal{D}_{\text{DPO}}^i \gets \{\}$ \Comment{\footnotesize Initialize DPO dataset}
    \For{\footnotesize each $(q, a^{\text{gold}}) \in \mathcal{Q}^i$}
        \State \footnotesize $a^{\text{gen}} \gets M_{\theta_{\text{SFT}}^i}(q)$; Add $(q, a^{\text{gold}}, a^{\text{gen}})$ to $\mathcal{D}_{\text{DPO}}^i$
    \EndFor

    \State \footnotesize $\Delta\theta_{\text{DPO}}^i \gets \underset{\Delta\theta}{\arg\min} \sum_{(q, a^{\text{gold}}, a^{\text{gen}}) \in \mathcal{D}_{\text{DPO}}^i} \mathcal{L}_{\text{DPO}}(M_{\theta_{\text{SFT}}^i+\Delta\theta}, q, a^{\text{gold}}, a^{\text{gen}})$
    \State \footnotesize $\theta^i \gets \theta_{\text{SFT}}^i + \Delta\theta_{\text{DPO}}^i$; $\text{acc}_{\text{DPO}}^i \gets \text{Evaluate}(M_{\theta^i}, \mathcal{Q}^i_{\text{iter\_sample}})$

    \State \footnotesize $\mathcal{Q}^i_{\text{eval}} \gets \text{GenerateQA}(\mathcal{C})$; $\mathcal{Q}^i_{\text{final\_sample}} \gets \text{SampleQA}(\mathcal{Q}^i_{\text{eval}}, S)$ 
    \State \footnotesize $\text{acc}_{\text{final}}^i \gets \text{Evaluate}(M_{\theta^i}, \mathcal{Q}^i_{\text{final\_sample}})$ \Comment{\footnotesize Evaluate generalization}

    \State \footnotesize $M_{\theta^i} \gets \text{SaveCheckpoint}(M_{\theta^i})$; $\mathcal{Q}^{i+1} \gets \mathcal{Q}^i_{\text{eval}}$; $\mathcal{Q}^{i+1}_{\text{iter\_sample}} \gets \mathcal{Q}^i_{\text{final\_sample}}$
\EndFor
\State \Return $M_{\theta^T}$
\end{algorithmic}
\end{algorithm}

\subsection{PASTA Algorithm}

The PASTA algorithm integrates new knowledge into the \btt{Target LLM} through $T$ iterations. The details of the PASTA algorithm are shown in Algorithm \ref{alg:main_process}.
Each iteration consists of the following three learning stages:

\noindent \textbf{1. CPT within PASTA Framework:} In the PASTA framework, the CPT stage performs autoregressive language modeling using the \btt{Augmented-Context} (extended context set $\mathcal{C}^{\text{para},i}$) \cite{radford2018improving}. This specialized pretraining approach allows for effective integration of new knowledge. The loss function $\mathcal{L}_{\text{CPT}}$ in PASTA is the cross-entropy of next token prediction:
\begin{align}
    \mathcal{L}_{\text{CPT}}(M_{\theta}, c^{\text{para}}) = -\frac{1}{|c^{\text{para}}|}\sum_{m=1}^{|c^{\text{para}}|} \log p_{\theta}(c^{\text{para}}_m | c^{\text{para}}_{<m}).
\end{align}

\noindent \textbf{2. SFT within PASTA Framework:} Following the CPT stage, the PASTA framework employs supervised fine-tuning with the \btt{Context-Derived QA} (question-answer set $\mathcal{Q}^i$) \cite{ouyang2022training} to adjust the model to generate appropriate answers to questions about the new knowledge. The SFT loss function in PASTA maximizes the generation probability of only the answer portion:
\begin{align}
    \mathcal{L}_{\text{SFT}}(M_{\theta}, q, a^{\text{gold}}) = -\frac{1}{|a^{\text{gold}}|}\sum_{m=1}^{|a^{\text{gold}}|} \log p_{\theta}(a^{\text{gold}}_m | q, a^{\text{gold}}_{<m}).
\end{align}

\noindent \textbf{3. DPO within PASTA Framework:} The final stage of the PASTA framework leverages DPO~\cite{NEURIPS2023_a85b405e} in a novel way to specifically address hallucination suppression in knowledge updating. While DPO was originally proposed as an efficient alternative to RLHF~\cite{christiano2017deep}, in PASTA it performs preference learning by contrasting model-generated answers with correct answers after SFT to suppress incorrect responses based on outdated knowledge. The PASTA DPO process creates triplets containing the question, gold answer, and model-generated answer, which are then used to optimize the model parameters as shown in Algorithm~\ref{alg:main_process}.

The DPO loss function within PASTA is defined using pairs of preferred and non-preferred answers:
\begin{align}
    \mathcal{L}_{\text{DPO}}(M_{\theta}, q, a^{\text{gold}}, a^{\text{gen}}) = -\log \sigma\left(\frac{1}{\beta}\left(\log\frac{p_{\theta}(a^{\text{gold}}|q)}{p_{\text{ref}}(a^{\text{gold}}|q)} - \log\frac{p_{\theta}(a^{\text{gen}}|q)}{p_{\text{ref}}(a^{\text{gen}}|q)}\right)\right).
\end{align}
Here, $p_{\text{ref}}$ is the reference model (the model after SFT), $\beta$ is a scaling parameter, and $\sigma$ is the sigmoid function.

\noindent Evaluation is conducted after each learning stage. Common samples $\mathcal{Q}^i_{\text{iter\_sample}}$ sampled from $\mathcal{Q}^i$ are used to measure changes within each iteration, and generalization performance is evaluated at the end of each iteration using $\mathcal{Q}^i_{\text{final\_sample}}$ sampled from newly generated $\mathcal{Q}^i_{\text{eval}}$.

\subsection{Prompt Instructions}
The prompt for \btt{Augmented-Context} generation (see Appendix~\ref{appendix:Augmented-Context_prompt}) provides instructions to expand \btt{context} data through paraphrasing by changing word order, style, expressions, and symbol representations while preserving proper nouns, numerical values, dates, and other specific information.
The prompt for \btt{Context-Derived QA} generation (see Appendix~\ref{appendix:prompt_qa_generation}) instructs that \Q should lead to a uniquely determined \A, ensuring that the \A is not ambiguous. It also directs the \btt{Generator LLM} to produce as many diverse generations as possible in a single run.

\section{Experiments}
In this research, we scrape Japanese news articles published after the knowledge cutoff of the \btt{Target LLM} from the web as new knowledge (\btt{context}) for training. That is, we perform knowledge updating on the \btt{Target LLM} with \btt{context} that the \btt{Target LLM} does not know, and evaluate how accurately it can answer questions about the \btt{context} that it did not know before training, thereby verifying the effectiveness of our proposed method.
Japanese, compared to English, often has more limited resources, which can make challenges in knowledge updating more apparent.
We vary the number of learning iterations and the generation volume of \btt{Augmented-Context} and \btt{Context-Derived QA} in our evaluation, aiming to gain insights into the architecture required to effectively integrate new knowledge into the \btt{Target LLM} and build a \btt{Target LLM} capable of accurately answering questions about that content.

\subsection{Context Dataset Construction}
We scraped Japanese online news articles published from May to July 2024, after the knowledge cutoff of the \btt{Target LLM}, and performed category classification (11 categories, see Appendix \ref{appendix:news_category_list} for details) and quality assessment.
From these, we selected the \btt{``Culture \& Entertainment''} category, which had a high number of articles and was suitable for knowledge updating verification. The reason is that information such as music rankings or celebrity appearance information tends to be frequently overwritten with new information daily. From this category, we randomly selected 128 high-quality articles (scoring 4 or higher on a 5-point scale) and summarized them to approximately 500 characters using an LLM if they exceeded that length.

\subsection{Experimental Setup}
\label{exp_setting}

In our experiments, we generated $P$ (=200) paraphrases and $Q$ (=50) question-answer pairs for each of the $K$ (=128) news articles (contexts). Examples of context expansion through paraphrasing are shown in Appendix \ref{appendix:paraphrase_exampl}, and examples of question-answer pair generation are shown in Figure~\ref{appendix:qa_genelation_exampl}. This provided $|\mathcal{C}^{\text{para},i}|$ (=$K \times P$ = 25,600) \btt{Augmented-Context} items and $|\mathcal{Q}^i|$ (=$K \times Q$ = 6,400) \btt{Context-Derived QA} pairs for training in each iteration. Training was optimized using LoRA~\cite{hu2021lora} (rank $r$ (=128)). The proposed method implemented $T$ (=7) iterations, and accuracy was measured by randomly sampling $S$ (=100) QA pairs at each stage (CPT, SFT, DPO) for evaluation. Within the same iteration, identical samples were used across all stages to track progress through the learning stages.

\paragraph{\btt{Target LLM}}
We adopted \emph{Llama-3.1-8B-Instruct} \cite{dubey2024llama} as the LLM (base model) for knowledge updating in this paper.
This is an 8B parameter model that was pre-trained and instruction-tuned for question answering with a knowledge cutoff as of December 2023 (prior to the news articles). Greedy decoding was used for answer generation for evaluation purposes and for generating data for DPO.
For other detailed experimental settings, please refer to Appendix \ref{appendix:parameter}.

\subsection{Evaluation Methods}

\paragraph{Answer Accuracy}
We implemented evaluation using the LLM-as-a-judge \cite{zheng2023judging} methodology at key points in the PASTA Framework. We evaluated after each stage (Initial, After CPT, After SFT, and After DPO), and at the end of each iteration by generating new \btt{Context-Derived QA} to assess generalization performance (Final evaluation).
As an evaluation metric, we define the accuracy of model $M_{\theta}$ with the following equation:

\begin{align}
\text{Acc}(M_{\theta}, \mathcal{Q}_{\text{sample}}, \mathcal{C}) = \frac{1}{|\mathcal{Q}_{\text{sample}}|} \sum_{q \in \mathcal{Q}_{\text{sample}}} \mathbb{I}[J(M_{\theta}(q), q, c_q) = 1]
\end{align}

This evaluation (where $J$ and $c_q$ are defined in Appendix \ref{appendix:eval_detail}) strictly judges the accuracy of proper nouns, numerical values, dates, and the absence of contradictions with the context content. Format differences (e.g., December 1, 2024 vs. 12/01/2024) are acceptable, but omissions or errors in information are considered incorrect. Details of the prompts used for evaluation are shown in Appendix~\ref{sec:prompt_llm_judge}, and details of the evaluation logic are shown in Appendix~\ref{appendix:eval_detail}.

\paragraph{Catastrophic Forgetting Evaluation of LLMs}
While the purpose of this paper is to build a specialized LLM capable of accurately answering questions about newly updated knowledge, we also evaluated the degree of catastrophic forgetting \cite{mccloskey1989catastrophic,french1999catastrophic} of the LLM's general capabilities using the Japanese MT-Bench++ benchmark \cite{uematsu2025japanese}.

\subsection{Results}

\begin{table}[t]
    \centering
    \caption{Accuracy comparison between the baseline method (SFT only) and the proposed method (CPT+SFT+DPO) across iterations. For After CPT, SFT, and DPO, evaluation samples from training data represent memorization ability, while Final evaluations use newly created \btt{Context-Derived QA} samples, representing generalization performance.}
    \label{tab:comparison_accuracy}
    \small
    \begin{tabular}{c|c|c|cccc|c}
        \toprule
        & \multicolumn{2}{c|}{\textbf{Baseline (SFT only)}} & \multicolumn{5}{c}{\textbf{Proposed Method (CPT+SFT+DPO)}} \\
        \cmidrule(lr){2-3} \cmidrule(lr){4-8}
        \textbf{Iter.} & \textbf{After SFT} & \textbf{Final} & \textbf{Initial} & \textbf{After CPT} & \textbf{After SFT} & \textbf{After DPO} & \textbf{Final} \\
        \midrule
        1 & 0.06 & 0.07 & 0.02 & 0.14 & 0.38 & 0.67 & 0.54 \\
        2 & 0.12 & 0.16 & -- & 0.52 & 0.43 & 0.68 & 0.65 \\
        3 & 0.16 & 0.13 & -- & 0.56 & 0.43 & 0.65 & 0.64 \\
        4 & 0.16 & 0.09 & -- & 0.65 & 0.49 & 0.67 & 0.66 \\
        5 & 0.22 & 0.19 & -- & 0.62 & 0.34 & 0.71 & \textbf{0.68} \\
        6 & 0.21 & 0.18 & -- & 0.65 & 0.47 & 0.63 & 0.67 \\
        7 & 0.24 & \textbf{0.26} & -- & 0.55 & 0.33 & 0.67 & 0.64 \\
        \bottomrule
    \end{tabular}
\end{table}

\paragraph{Accuracy Overview of the Proposed Method}
Through our experiments, the effectiveness of the PASTA framework became evident. The \btt{Target LLM} before knowledge updating showed only 0.02 accuracy when answering questions about news articles published after its knowledge cutoff. After training with our framework, we observed substantial improvement. Under optimal conditions with increased \btt{Context-Derived QA} volume, determined through parameter sensitivity analysis, we achieved \btt{a maximum accuracy of 0.82}. Below, we present results from our basic parameter settings followed by detailed analyses.

\paragraph{Results with Basic Experimental Settings}
Table~\ref{tab:comparison_accuracy} shows a comparison between our proposed method with basic parameter settings (\btt{Context-Derived QA} multiplier of 50) and the baseline (SFT only). In the proposed method, accuracy improved progressively through each learning stage (CPT, SFT, DPO) in the first iteration, reaching 0.67 after DPO. The Final metric, which evaluates generalization performance, recorded 0.54. From the second iteration onward, performance improved steadily, with the Final metric reaching 0.65 and then stabilizing in the upper 0.6 range in subsequent iterations. The highest Final metric value among the 7 iterations was 0.68.
In contrast, the baseline trained with SFT only reached only 0.26 accuracy even after 7 iterations. The proposed method achieved 0.54 accuracy in just the first iteration and 0.65 by the second iteration, demonstrating that this difference cannot be explained merely by the number of learning steps. This provides compelling evidence for the effectiveness of the integrated CPT, SFT, and DPO framework.
Note that evaluation was conducted by extracting 100 random samples from each \btt{Context-Derived QA}, so measurements contain some statistical variation. This experimental design prioritizes efficiently conducting multiple experiments under various parameter conditions to identify the characteristics of effective knowledge updating methods.

\paragraph{Sensitivity Analysis of \btt{Augmented-Context} and \btt{Context-Derived QA}}
We analyzed the effect of increasing and decreasing the volume of \btt{Augmented-Context} and \btt{Context-Derived QA} data to determine which data type has a greater impact on accuracy. Table~\ref{tab:reduced_data_comparison} shows that \btt{Context-Derived QA} has a more significant effect on improving accuracy.

\begin{table}[t]
    \centering
    \caption{Accuracy comparison with 1/10 data reduction: \btt{Augmented-Context} reduction (left) vs. \btt{Context-Derived QA} reduction (right). Final represents generalization performance evaluation.}
    \label{tab:reduced_data_comparison}
    \footnotesize
    \begin{tabular}{c|ccc|c|ccc|c}
        \toprule
        & \multicolumn{4}{c|}{\textbf{1/10 Augmented-Context}} & \multicolumn{4}{c}{\textbf{1/10 Context-Derived QA}} \\
        \cmidrule(lr){2-5} \cmidrule(lr){6-9}
        \textbf{Iter.} & \textbf{CPT} & \textbf{SFT} & \textbf{DPO} & \textbf{Final} & \textbf{CPT} & \textbf{SFT} & \textbf{DPO} & \textbf{Final} \\
        \midrule
        1 & 0.03 & 0.17 & 0.31 & 0.32 & 0.12 & 0.23 & 0.26 & 0.20 \\
        2 & 0.31 & 0.27 & 0.55 & 0.42 & 0.26 & 0.31 & 0.34 & 0.26 \\
        3 & 0.45 & 0.25 & 0.50 & 0.42 & 0.30 & 0.28 & 0.39 & 0.28 \\
        4 & 0.51 & 0.33 & 0.59 & \textbf{0.63} & 0.29 & 0.42 & 0.45 & 0.40 \\
        5 & 0.51 & 0.37 & 0.60 & 0.58 & 0.45 & 0.38 & 0.47 & 0.29 \\
        6 & 0.57 & 0.46 & 0.62 & 0.61 & 0.36 & 0.42 & 0.42 & \textbf{0.49} \\
        7 & 0.67 & 0.47 & 0.64 & 0.58 & 0.44 & 0.47 & 0.54 & 0.39 \\
        8 & 0.65 & 0.54 & 0.67 & 0.55 & 0.51 & 0.50 & 0.54 & 0.37 \\
        \bottomrule
    \end{tabular}
\end{table}

\paragraph{Increasing \btt{Context-Derived QA} by 10 times}
Based on the above results, we experimented with increasing the \btt{Context-Derived QA}—which was shown to be more effective for accuracy improvement—to 10 times that of the proposed method to determine how much accuracy would improve. The results are shown in Table~\ref{tab:qa_volume_comparison}.
This achieved higher accuracy than the proposed method, reaching 0.82 accuracy in the Final evaluation of the 2nd iteration. With 10 times the proposed method, the \btt{Context-Derived QA} has 500 pairs per context.

\begin{table}[h]
    \centering
    \caption{Impact of \btt{Context-Derived QA} volume on accuracy: Comparison of 1/10×, 1×, and 10× the proposed method. Final represents generalization performance evaluation.}
    \label{tab:qa_volume_comparison}
    \footnotesize
    \setlength{\tabcolsep}{3.8pt}
    \begin{tabular}{c|ccc|c|ccc|c|ccc|c}
        \toprule
        & \multicolumn{4}{c|}{\textbf{1/10× QA Volume}} & \multicolumn{4}{c|}{\textbf{1× QA Volume (Proposed)}} & \multicolumn{4}{c}{\textbf{10× QA Volume}} \\
        \cmidrule(lr){2-5} \cmidrule(lr){6-9} \cmidrule(lr){10-13}
        \textbf{Iter.} & \textbf{CPT} & \textbf{SFT} & \textbf{DPO} & \textbf{Final} & \textbf{CPT} & \textbf{SFT} & \textbf{DPO} & \textbf{Final} & \textbf{CPT} & \textbf{SFT} & \textbf{DPO} & \textbf{Final} \\
        \midrule
        1 & 0.12 & 0.23 & 0.26 & 0.20 & 0.14 & 0.38 & 0.67 & 0.54 & 0.10 & 0.29 & 0.79 & 0.74 \\
        2 & 0.26 & 0.31 & 0.34 & \textbf{0.26} & 0.52 & 0.43 & 0.68 & \textbf{0.65} & 0.68 & 0.59 & 0.91 & \textbf{0.82} \\
        \bottomrule
    \end{tabular}
\end{table}

\paragraph{Differences in Accuracy by Training Volume per Iteration}
Table~\ref{tab:training_methods_comparison} shows a comparison of reducing the data volume of \btt{Augmented-Context} and \btt{Context-Derived QA} per iteration to 1/10 of the proposed method and training with more iterations (called the Small Batch Method).
Since 1 iteration of the proposed method and 10 iterations of the Small Batch Method have learned the same amount of data, they are compared side by side in the table.
We found that the Final accuracy of the 4th iteration of the proposed method and the Final accuracy of the Small Batch Method are both 0.66, indicating that similar accuracy can be achieved.

\begin{table}[H]
    \centering
    \caption{Comparison of the proposed method and small batch method with equivalent total training data volume. Final represents generalization performance evaluation.}
    \label{tab:training_methods_comparison}
    \small
    \setlength{\tabcolsep}{3.8pt}
    \begin{tabular}{c|ccc|c|c|ccc|c}
        \toprule
        \multicolumn{5}{c|}{\textbf{Proposed Method}} & \multicolumn{5}{c}{\textbf{Small Batch Method}} \\
        \cmidrule(lr){1-5} \cmidrule(lr){6-10}
        \textbf{Iter.} & \textbf{CPT} & \textbf{SFT} & \textbf{DPO} & \textbf{Final} & \textbf{Equiv.} & \textbf{CPT} & \textbf{SFT} & \textbf{DPO} & \textbf{Final} \\
        \midrule
        1 & 0.14 & 0.38 & 0.67 & 0.54 & 10 & 0.42 & 0.43 & 0.41 & 0.29 \\
        2 & 0.52 & 0.43 & 0.68 & 0.65 & 20 & 0.41 & 0.45 & 0.44 & 0.40 \\
        3 & 0.56 & 0.43 & 0.65 & 0.64 & 30 & 0.39 & 0.41 & 0.54 & 0.50 \\
        4 & 0.65 & 0.49 & 0.67 & \textbf{0.66} & 40 & 0.52 & 0.53 & 0.60 & \textbf{0.66} \\
        \bottomrule
    \end{tabular}
\end{table}

\paragraph{Ablation: Accuracy Differences by Presence/Absence of CPT, SFT, DPO}
We conducted experiments to observe changes in accuracy when omitting one of the CPT, SFT, or DPO learning processes from the proposed method, and show the results in Table~\ref{tab:ablation_study}. These results confirm that the overall framework of the proposed method enhances accuracy, and omitting any one of the learning processes significantly reduces accuracy.

\begin{table}[h]
    \centering
    \caption{Ablation study: Comparison of the proposed method and variant methods missing certain components. Final represents generalization performance evaluation.}
    \label{tab:ablation_study}
    \setlength{\tabcolsep}{4pt}
    \small
    \begin{tabular}{c|ccc|c|cc|c|cc|c|cc|c}
        \toprule
        & \multicolumn{4}{c|}{\textbf{Proposed Method}} & \multicolumn{3}{c|}{\textbf{w/o CPT}} & \multicolumn{3}{c|}{\textbf{w/o SFT}} & \multicolumn{3}{c}{\textbf{w/o DPO}} \\
        \cmidrule(lr){2-5} \cmidrule(lr){6-8} \cmidrule(lr){9-11} \cmidrule(lr){12-14}
        \textbf{Iter} & \textbf{CPT} & \textbf{SFT} & \textbf{DPO} & \textbf{Final} & \textbf{SFT} & \textbf{DPO} & \textbf{Final} & \textbf{CPT} & \textbf{DPO} & \textbf{Final} & \textbf{CPT} & \textbf{SFT} & \textbf{Final} \\
        \midrule
        1 & 0.14 & 0.38 & 0.67 & 0.54 & 0.08 & 0.15 & 0.09 & 0.18 & 0.26 & \textbf{0.30} & 0.08 & 0.25 & 0.25 \\
        2 & 0.52 & 0.43 & 0.68 & 0.65 & 0.19 & 0.25 & 0.22 & 0.23 & 0.23 & 0.17 & 0.28 & 0.33 & 0.40 \\
        3 & 0.56 & 0.43 & 0.65 & 0.64 & 0.30 & 0.29 & 0.29 & 0.25 & 0.22 & 0.18 & 0.35 & 0.44 & 0.34 \\
        4 & 0.65 & 0.49 & 0.67 & 0.66 & 0.14 & 0.28 & \textbf{0.32} & 0.19 & 0.18 & 0.20 & 0.50 & 0.53 & \textbf{0.46} \\
        5 & 0.62 & 0.34 & 0.71 & \textbf{0.68} & 0.15 & 0.20 & 0.15 & 0.27 & 0.13 & 0.22 & 0.44 & 0.46 & 0.31 \\
        6 & 0.65 & 0.47 & 0.63 & 0.67 & 0.16 & 0.31 & 0.29 & 0.25 & 0.15 & 0.16 & 0.52 & 0.38 & 0.35 \\
        \bottomrule
    \end{tabular}
\end{table}

\paragraph{Qualitative Evaluation}
Figure~\ref{fig:qualitative_results_fujii} shows an example of an actual news article used in the experiment and output examples from each model. The process of gradually being able to produce concise and correct answers through DPO, which is the knowledge updating hypothesis method based on the self-training approach in the proposed method, can be confirmed. Other qualitative evaluations are included in Appendix~\ref{appendix:eval_document}.

\begin{figure}[H]
\centering
\begin{minipage}{\linewidth}
    \centering
    \begin{tcolorbox}[
      colback=gray!5,
      colframe=black,
      arc=1.5mm,
      boxrule=0.5pt,
      width=0.98\linewidth,
      left=1mm,
      right=1mm,
      top=0.5mm,
      bottom=0.5mm,
      title={Qualitative Evaluation: Response Evolution Across Learning Stages (Iteration 1)},
      titlerule=0.5pt,
      colbacktitle=gray!20,
      coltitle=black,
      fonttitle=\footnotesize\bfseries
    ]
    \scriptsize
    \colorbox{blue!10}{\tiny\textbf{Context}} \\[-1pt]
    On Billboard JAPAN's "Japan Songs (Country Chart)," Creepy Nuts' "Bling-Bang-Bang-Born" reached \#1 in six countries including the United States, United Kingdom, France, Brazil, South Africa, and Singapore during the April 19-25, 2024 tabulation period. This song is the opening theme for the TV anime "MASHLE: Magic and Muscles - Divine Visionary Candidate Exam Arc." Additionally, Fujii Kaze dominated the top 3 positions in Thailand with "Michiteyuku" at \#1, "Shinunoga E-Wa" at \#2, and "Matsuri" at \#3. Furthermore, six songs by Fujii Kaze ranked in Thailand's top 20.

    \vspace{0.3em}
    \colorbox{blue!10}{\tiny\textbf{Question}} \\[-1pt]
    Please tell me about the three songs by Fujii Kaze that dominated the top 3 on the Billboard JAPAN country chart from April 19-25, 2024.

    \vspace{0.3em}
    \colorbox{red!15}{\tiny\textbf{Response after CPT} \textit{[LLM as a judge: Incorrect]}} \\[-1pt]
    During the April 19-25, 2024 period on Billboard JAPAN's "Japan Songs" country chart, \textcolor{red}{Fujii Kaze topped charts in six countries including the United States, United Kingdom, France, Brazil, South Africa, and more. His track "Michiteyuku" reached \#16 in the US, \#17 in the UK, and \#6 in France. "Shinunoga E-Wa" ranked \#93 in the US, \#76 in the UK, and \#33 in France. "Matsuri" placed \#93 in the US, \#76 in the UK, and \#46 in France.} Fujii Kaze's music is gaining attention worldwide.

    \vspace{0.3em}
    \colorbox{red!15}{\tiny\textbf{Response after SFT} \textit{[LLM as a judge: Incorrect]}} \\[-1pt]
    Fujii Kaze's songs "Michiteyuku" ranked \#1, "Shinunoga E-Wa" ranked \#2, and "Matsuri" ranked \#3 in Thailand, dominating the top 3 positions. \textcolor{red}{Additionally, "Matsuri" has been \#1 on Thailand's chart for 6 consecutive weeks.} His songs are very popular on Thailand's charts. This data was collected during the April 19-25, 2024 period.

    \vspace{0.3em}
    \colorbox{green!15}{\tiny\textbf{Response after DPO} \textit{[LLM as a judge: Correct]}} \\[-1pt]
    Fujii Kaze's songs "Michiteyuku" ranked \#1, "Shinunoga E-Wa" ranked \#2, and "Matsuri" ranked \#3 in Thailand.

    \end{tcolorbox}
    \vspace{-6pt}
    \footnotesize\caption{Example of response quality changes across learning stages. \textcolor{red}{Hallucinations (red text)}.}
    \label{fig:qualitative_results_fujii}
\end{minipage}
\end{figure}

\subsection{Degree of Catastrophic Forgetting Evaluation}

While the purpose of this paper is to build a specialized LLM capable of accurately answering questions about newly updated knowledge, we also evaluated the degree of catastrophic forgetting \cite{mccloskey1989catastrophic,french1999catastrophic} of the LLM's general capabilities using the Japanese MT-Bench++ benchmark \cite{uematsu2025japanese}.
Table~\ref{tab:jp_mt_bench_pp_results} shows the evaluation results of the \btt{Target LLM} using the proposed method.
As shown in Table~\ref{tab:jp_mt_bench_pp_results}, the \btt{Target LLM} shows a slight decrease in accuracy compared to the base model as the learning iterations progress in the ``All'' category, but it sometimes exceeds the base model in ``Reasoning'' and ``Writing.'' We confirmed that the degree of degradation in the general language processing capabilities of the \btt{Target LLM} accompanying knowledge updating is minimal, and catastrophic forgetting does not occur.
We compare the scores of the \btt{Target LLM} before training with those of the proposed method.

\begin{table}[t]
    \centering
    \caption{Performance comparison on Japanese MT-bench++ benchmark.}
    \label{tab:jp_mt_bench_pp_results}
    \small
    \setlength{\tabcolsep}{4pt}
    \begin{tabular}{l|ccccc|c}
        \toprule
        \textbf{Model} & \textbf{Humanities} & \textbf{Reasoning} & \textbf{Roleplay} & \textbf{STEM} & \textbf{Writing} & \textbf{All} \\
        \midrule
        Llama-3.1-8B-Instruct & \textbf{4.72} & 4.17 & \textbf{4.33} & \textbf{4.50} & 4.38 & \textbf{4.42} \\
        \midrule
        \multicolumn{7}{l}{\textbf{Proposed Model}} \\
        \cmidrule{1-7}
        Iteration 1 & 3.97 & \textbf{4.43} & 4.20 & 4.00 & 3.93 & 4.11 \\
        Iteration 2 & 4.17 & \textbf{4.43} & 3.80 & 4.33 & 4.33 & 4.21 \\
        Iteration 3 & 3.77 & 3.70 & 3.67 & 3.77 & 4.47 & 3.87 \\
        Iteration 4 & 3.63 & 3.37 & 3.53 & 4.10 & 4.40 & 3.81 \\
        Iteration 5 & 3.63 & 3.60 & 3.63 & 3.80 & 4.43 & 3.82 \\
        Iteration 6 & 3.93 & 3.70 & 4.00 & 3.70 & 4.27 & 3.92 \\
        Iteration 7 & 3.60 & 3.23 & 3.87 & 3.70 & \textbf{4.67} & 3.81 \\
        \bottomrule
    \end{tabular}
\end{table}

\section{Conclusion}
In this paper, we demonstrated that our simple yet powerful PASTA framework significantly enhances knowledge updating capabilities, achieving \btt{0.82 accuracy} with optimized configurations—a remarkable improvement from the initial 0.02 accuracy. The primary goal of PASTA is to build specialized LLMs that can accurately respond to questions about newly acquired knowledge, which we successfully accomplished. PASTA outperforms conventional approaches through systematic integration of established techniques. Sensitivity analysis revealed that \btt{Context-Derived QA} volume primarily drives performance gains, while PASTA's novel self-learning DPO process effectively enables knowledge overwriting and suppresses hallucinations simultaneously. While not the primary focus, evaluation using Japanese MT-Bench++ showed that the general language capabilities remained largely intact with minimal degradation. These findings establish PASTA as an effective framework for creating knowledge-specialized LLMs for specific domains. Future work includes application to diverse domains, evaluation refinements, and mechanisms for handling interdependent knowledge updates.

\section{Limitations}
This research has several limitations. First, experiments were conducted using only a single pre-trained LLM and Japanese news articles from a specific genre, limiting the demonstrated generalizability of the method. Computational resource constraints prevented extensive exploration of iteration counts and hyperparameters, though more optimal settings may exist. The model developed in this study is specialized for question answering about new knowledge rather than improving comprehensive LLM capabilities. The evaluation methodology using LLM as a judge introduces potential bias (\citealt{zheng2023judging}). Furthermore, our comparative experiments did not rigorously equalize computational resources across different methods, which remains a future research challenge. Considering these limitations, future research should investigate these approaches with diverse LLMs, datasets, and more rigorous experimental protocols.

\section*{Acknowledgments}
This study was carried out using the TSUBAME4.0 supercomputer at Institute of Science Tokyo.

%\newpage
{\small
\bibliographystyle{plainnat}
\bibliography{custom}

\begin{thebibliography}{27}
\providecommand{\natexlab}[1]{#1}
\providecommand{\url}[1]{\texttt{#1}}
\expandafter\ifx\csname urlstyle\endcsname\relax
  \providecommand{\doi}[1]{doi: #1}\else
  \providecommand{\doi}{doi: \begingroup \urlstyle{rm}\Url}\fi

\bibitem[Allen-Zhu and Li(2023)]{allen2023physics}
Zeyuan Allen-Zhu and Yuanzhi Li.
\newblock Physics of language models: Part 3.1, knowledge storage and extraction.
\newblock \emph{arXiv preprint arXiv:2309.14316}, 2023.

\bibitem[Brown et~al.(2024)Brown, Juravsky, Ehrlich, Clark, Le, R{\'e}, and Mirhoseini]{brown2024large}
Bradley Brown, Jordan Juravsky, Ryan Ehrlich, Ronald Clark, Quoc~V Le, Christopher R{\'e}, and Azalia Mirhoseini.
\newblock Large language monkeys: Scaling inference compute with repeated sampling.
\newblock \emph{arXiv preprint arXiv:2407.21787}, 2024.

\bibitem[Christiano et~al.(2017)Christiano, Leike, Brown, Martic, Legg, and Amodei]{christiano2017deep}
Paul~F Christiano, Jan Leike, Tom Brown, Miljan Martic, Shane Legg, and Dario Amodei.
\newblock Deep reinforcement learning from human preferences.
\newblock \emph{Advances in neural information processing systems}, 30, 2017.

\bibitem[Dubey et~al.(2024)Dubey, Jauhri, Pandey, Kadian, Al-Dahle, Letman, Mathur, Schelten, Yang, Fan, et~al.]{dubey2024llama}
Abhimanyu Dubey, Abhinav Jauhri, Abhinav Pandey, Abhishek Kadian, Ahmad Al-Dahle, Aiesha Letman, Akhil Mathur, Alan Schelten, Amy Yang, Angela Fan, et~al.
\newblock The llama 3 herd of models.
\newblock \emph{arXiv preprint arXiv:2407.21783}, 2024.

\bibitem[Farquhar et~al.(2024)Farquhar, Kossen, Kuhn, and Gal]{farquhar2024detecting}
Sebastian Farquhar, Jannik Kossen, Lorenz Kuhn, and Yarin Gal.
\newblock Detecting hallucinations in large language models using semantic entropy.
\newblock \emph{Nature}, 630\penalty0 (8017):\penalty0 625--630, 2024.

\bibitem[Frankle and Carbin(2018)]{frankle2018lottery}
Jonathan Frankle and Michael Carbin.
\newblock The lottery ticket hypothesis: Finding sparse, trainable neural networks.
\newblock \emph{arXiv preprint arXiv:1803.03635}, 2018.

\bibitem[French(1999)]{french1999catastrophic}
Robert~M French.
\newblock Catastrophic forgetting in connectionist networks.
\newblock \emph{Trends in cognitive sciences}, 3\penalty0 (4):\penalty0 128--135, 1999.

\bibitem[Han et~al.(2015)Han, Pool, Tran, and Dally]{han2015learning}
Song Han, Jeff Pool, John Tran, and William Dally.
\newblock Learning both weights and connections for efficient neural network.
\newblock \emph{Advances in neural information processing systems}, 28, 2015.

\bibitem[Hase et~al.(2024)Hase, Hofweber, Zhou, Stengel-Eskin, and Bansal]{hase2024fundamental}
Peter Hase, Thomas Hofweber, Xiang Zhou, Elias Stengel-Eskin, and Mohit Bansal.
\newblock Fundamental problems with model editing: How should rational belief revision work in llms?
\newblock \emph{arXiv preprint arXiv:2406.19354}, 2024.

\bibitem[Hu et~al.(2021)Hu, Shen, Wallis, Allen-Zhu, Li, Wang, Wang, and Chen]{hu2021lora}
Edward~J Hu, Yelong Shen, Phillip Wallis, Zeyuan Allen-Zhu, Yuanzhi Li, Shean Wang, Lu~Wang, and Weizhu Chen.
\newblock Lora: Low-rank adaptation of large language models.
\newblock \emph{arXiv preprint arXiv:2106.09685}, 2021.

\bibitem[Kuroki et~al.(2024)Kuroki, Nakamura, Akiba, and Tang]{kuroki2024agent}
So~Kuroki, Taishi Nakamura, Takuya Akiba, and Yujin Tang.
\newblock Agent skill acquisition for large language models via cycleqd.
\newblock \emph{arXiv preprint arXiv:2410.14735}, 2024.

\bibitem[Lee et~al.(2024)Lee, Phatale, Mansoor, Lu, Mesnard, Ferret, Bishop, Hall, Carbune, and Rastogi]{lee2024rlaif}
Harrison Lee, Samrat Phatale, Hassan Mansoor, Kellie~Ren Lu, Thomas Mesnard, Johan Ferret, Colton Bishop, Ethan Hall, Victor Carbune, and Abhinav Rastogi.
\newblock {RLAIF}: Scaling reinforcement learning from human feedback with {AI} feedback, 2024.
\newblock URL \url{https://openreview.net/forum?id=AAxIs3D2ZZ}.

\bibitem[Lewis et~al.(2020)Lewis, Perez, Piktus, Petroni, Karpukhin, Goyal, K{\"u}ttler, Lewis, Yih, Rockt{\"a}schel, et~al.]{lewis2020retrieval}
Patrick Lewis, Ethan Perez, Aleksandra Piktus, Fabio Petroni, Vladimir Karpukhin, Naman Goyal, Heinrich K{\"u}ttler, Mike Lewis, Wen-tau Yih, Tim Rockt{\"a}schel, et~al.
\newblock Retrieval-augmented generation for knowledge-intensive nlp tasks.
\newblock \emph{Advances in Neural Information Processing Systems}, 33:\penalty0 9459--9474, 2020.

\bibitem[Li et~al.(2023)Li, Yu, Zhou, Schick, Levy, Zettlemoyer, Weston, and Lewis]{li2023self}
Xian Li, Ping Yu, Chunting Zhou, Timo Schick, Omer Levy, Luke Zettlemoyer, Jason Weston, and Mike Lewis.
\newblock Self-alignment with instruction backtranslation.
\newblock \emph{arXiv preprint arXiv:2308.06259}, 2023.

\bibitem[Madaan et~al.(2024)Madaan, Tandon, Gupta, Hallinan, Gao, Wiegreffe, Alon, Dziri, Prabhumoye, Yang, et~al.]{madaan2024self}
Aman Madaan, Niket Tandon, Prakhar Gupta, Skyler Hallinan, Luyu Gao, Sarah Wiegreffe, Uri Alon, Nouha Dziri, Shrimai Prabhumoye, Yiming Yang, et~al.
\newblock Self-refine: Iterative refinement with self-feedback.
\newblock \emph{Advances in Neural Information Processing Systems}, 36, 2024.

\bibitem[McCloskey and Cohen(1989)]{mccloskey1989catastrophic}
Michael McCloskey and Neal~J Cohen.
\newblock Catastrophic interference in connectionist networks: The sequential learning problem.
\newblock In \emph{Psychology of learning and motivation}, volume~24, pages 109--165. Elsevier, 1989.

\bibitem[Ouyang et~al.(2022)Ouyang, Wu, Jiang, Almeida, Wainwright, Mishkin, Zhang, Agarwal, Slama, Ray, et~al.]{ouyang2022training}
Long Ouyang, Jeffrey Wu, Xu~Jiang, Diogo Almeida, Carroll Wainwright, Pamela Mishkin, Chong Zhang, Sandhini Agarwal, Katarina Slama, Alex Ray, et~al.
\newblock Training language models to follow instructions with human feedback.
\newblock \emph{Advances in neural information processing systems}, 35:\penalty0 27730--27744, 2022.

\bibitem[Radford(2018)]{radford2018improving}
Alec Radford.
\newblock Improving language understanding by generative pre-training.
\newblock 2018.

\bibitem[Radford et~al.(2019)Radford, Wu, Child, Luan, Amodei, Sutskever, et~al.]{radford2019language}
Alec Radford, Jeffrey Wu, Rewon Child, David Luan, Dario Amodei, Ilya Sutskever, et~al.
\newblock Language models are unsupervised multitask learners.
\newblock \emph{OpenAI blog}, 1\penalty0 (8):\penalty0 9, 2019.

\bibitem[Rafailov et~al.(2023)Rafailov, Sharma, Mitchell, Manning, Ermon, and Finn]{NEURIPS2023_a85b405e}
Rafael Rafailov, Archit Sharma, Eric Mitchell, Christopher~D Manning, Stefano Ermon, and Chelsea Finn.
\newblock Direct preference optimization: Your language model is secretly a reward model.
\newblock In A.~Oh, T.~Naumann, A.~Globerson, K.~Saenko, M.~Hardt, and S.~Levine, editors, \emph{Advances in Neural Information Processing Systems}, volume~36, pages 53728--53741. Curran Associates, Inc., 2023.
\newblock URL \url{https://proceedings.neurips.cc/paper_files/paper/2023/file/a85b405ed65c6477a4fe8302b5e06ce7-Paper-Conference.pdf}.

\bibitem[Roziere et~al.(2023)Roziere, Gehring, Gloeckle, Sootla, Gat, Tan, Adi, Liu, Sauvestre, Remez, et~al.]{roziere2023code}
Baptiste Roziere, Jonas Gehring, Fabian Gloeckle, Sten Sootla, Itai Gat, Xiaoqing~Ellen Tan, Yossi Adi, Jingyu Liu, Romain Sauvestre, Tal Remez, et~al.
\newblock Code llama: Open foundation models for code.
\newblock \emph{arXiv preprint arXiv:2308.12950}, 2023.

\bibitem[Uematsu et~al.(2025)Uematsu, Fukuda, Kawahara, and Shibata]{uematsu2025japanese}
Takuya Uematsu, So~Fukuda, Daisuke Kawahara, and Tomohide Shibata.
\newblock {Japanese MT-bench++: A Large-Scale Japanese Benchmark with More Natural Multi-turn Dialogue Settings}.
\newblock In \emph{Proceedings of the 31st Annual Conference of the Association for Natural Language Processing}, pages 3569--3574. Association for Natural Language Processing, Japan, March 2025.
\newblock URL \url{https://www.anlp.jp/proceedings/annual_meeting/2025/pdf_dir/D9-1.pdf}.
\newblock Paper written in Japanese.

\bibitem[Wang et~al.(2024{\natexlab{a}})Wang, Shi, Li, Li, Dong, Zhang, Jiao, and Mei]{wang2024languagemodelscollapsetrained}
Lecheng Wang, Xianjie Shi, Ge~Li, Jia Li, Yihong Dong, Xuanming Zhang, Wenpin Jiao, and Hong Mei.
\newblock Why language models collapse when trained on recursively generated text, 2024{\natexlab{a}}.
\newblock URL \url{https://arxiv.org/abs/2412.14872}.

\bibitem[Wang et~al.(2024{\natexlab{b}})Wang, Kulikov, Golovneva, Yu, Yuan, Dwivedi-Yu, Pang, Fazel-Zarandi, Weston, and Li]{wang2024self}
Tianlu Wang, Ilia Kulikov, Olga Golovneva, Ping Yu, Weizhe Yuan, Jane Dwivedi-Yu, Richard~Yuanzhe Pang, Maryam Fazel-Zarandi, Jason Weston, and Xian Li.
\newblock Self-taught evaluators.
\newblock \emph{arXiv preprint arXiv:2408.02666}, 2024{\natexlab{b}}.

\bibitem[Wei et~al.(2024)Wei, Karina, Chung, Jiao, Papay, Glaese, Schulman, and Fedus]{wei2024measuring}
Jason Wei, Nguyen Karina, Hyung~Won Chung, Yunxin~Joy Jiao, Spencer Papay, Amelia Glaese, John Schulman, and William Fedus.
\newblock Measuring short-form factuality in large language models.
\newblock \emph{arXiv preprint arXiv:2411.04368}, 2024.

\bibitem[Yuan et~al.(2024)Yuan, Pang, Cho, Sukhbaatar, Xu, and Weston]{yuan2024self}
Weizhe Yuan, Richard~Yuanzhe Pang, Kyunghyun Cho, Sainbayar Sukhbaatar, Jing Xu, and Jason Weston.
\newblock Self-rewarding language models.
\newblock \emph{arXiv preprint arXiv:2401.10020}, 2024.

\bibitem[Zheng et~al.(2023)Zheng, Chiang, Sheng, Zhuang, Wu, Zhuang, Lin, Li, Li, Xing, et~al.]{zheng2023judging}
Lianmin Zheng, Wei-Lin Chiang, Ying Sheng, Siyuan Zhuang, Zhanghao Wu, Yonghao Zhuang, Zi~Lin, Zhuohan Li, Dacheng Li, Eric Xing, et~al.
\newblock Judging llm-as-a-judge with mt-bench and chatbot arena.
\newblock \emph{Advances in Neural Information Processing Systems}, 36:\penalty0 46595--46623, 2023.

\end{thebibliography}
}

\appendix

\section{Evaluation Methodology Details}
\label{appendix:eval_detail}
As an evaluation metric, we define the accuracy of model $M_{\theta}$ using the following equation:

\begin{align}
\text{Acc}(M_{\theta}, \mathcal{Q}_{\text{sample}}, \mathcal{C}) = \frac{1}{|\mathcal{Q}_{\text{sample}}|} \sum_{q \in \mathcal{Q}_{\text{sample}}} \mathbb{I}[J(M_{\theta}(q), q, c_q) = 1]
\end{align}

Where:
\begin{itemize}
\item $\mathcal{Q}_{\text{sample}}$ is the set of questions sampled for evaluation ($|\mathcal{Q}_{\text{sample}}| = S = 100$)
\item $c_q$ is the news article context corresponding to question $q$
\item $M_{\theta}(q)$ is the answer generated by the model for question $q$
\item $\mathbb{I}[\cdot]$ is the indicator function
\item $J(a^{\text{gen}}, q, c)$ is the LLM as a judge judgment function
\end{itemize}

The judge function $J(a^{\text{gen}}, q, c)$ returns 1 (correct) if the following conditions are met, and 0 (incorrect) otherwise (the actual implementation uses the prompt described in Appendix~\ref{sec:prompt_llm_judge}):
\begin{align}
J(a^{\text{gen}}, q, c) = 
\begin{cases}
1, & \text{if } a^{\text{gen}} \text{ is factually accurate and consistent with context } c \text{ for question } q \\
0, & \text{otherwise}
\end{cases}
\end{align}

Evaluations at each learning stage are defined as follows:
\begin{align}
\text{acc}_{\text{CPT}}^i &= \text{Acc}(M_{\theta_{\text{CPT}}^i}, \mathcal{Q}^i_{\text{iter\_sample}}, \mathcal{C}) \\
\text{acc}_{\text{SFT}}^i &= \text{Acc}(M_{\theta_{\text{SFT}}^i}, \mathcal{Q}^i_{\text{iter\_sample}}, \mathcal{C}) \\
\text{acc}_{\text{DPO}}^i &= \text{Acc}(M_{\theta^i}, \mathcal{Q}^i_{\text{iter\_sample}}, \mathcal{C}) \\
\text{acc}_{\text{final}}^i &= \text{Acc}(M_{\theta^i}, \mathcal{Q}^i_{\text{final\_sample}}, \mathcal{C})
\end{align}

Evaluation samples are generated by random sampling as follows:
\begin{align}
\mathcal{Q}^i_{\text{iter\_sample}} &= \text{SampleQA}(\mathcal{Q}^i, S) \\
\mathcal{Q}^i_{\text{final\_sample}} &= \text{SampleQA}(\mathcal{Q}^i_{\text{eval}}, S)
\end{align}

Where $S=100$ is the sample size, and $\text{SampleQA}(\mathcal{Q}, S)$ is a function that randomly extracts only the question portion from the question-answer pair set $\mathcal{Q}$ in $S$ quantities.

This evaluation strictly judges the accuracy of proper nouns, numerical values, dates, and the absence of contradictions with the context content. Format differences (e.g., December 1, 2024 vs. 12/01/2024) are acceptable, but omissions or errors in information are considered incorrect. Details of the prompts used for evaluation are shown in Appendix~\ref{sec:prompt_llm_judge}.

\section{Details of the PASTA Algorithm}
\label{appendix:full_algorithm}

The details of the data generation process part of the PASTA algorithm that could not be fully described in the main text are shown in Algorithm \ref{alg:data_generation}, and the auxiliary functions are shown in Algorithm \ref{alg:helper_functions}.

\begin{algorithm}
\caption{PASTA: Initialization and Data Generation Process}
\label{alg:data_generation}
\begin{algorithmic}[1]
\State \textbf{Input:} News article set $\mathcal{C} = \{c_1, c_2, \ldots, c_K\}$, \btt{Generator LLM} $G$, initial \btt{Target LLM} $M_{\theta^0}$, maximum number of iterations $T$, number of paraphrases $P$, number of QA pairs $Q$, number of evaluation samples $S$
\State \textbf{Output:} Trained \btt{Target LLM} $M_{\theta^T}$
\State // Initial evaluation
\State $\mathcal{Q}^0_{\text{eval}} \gets \text{GenerateQA}(\mathcal{C})$ \Comment{Generate QA for initial evaluation}
\State $\mathcal{Q}^0_{\text{sample}} \gets \text{SampleQA}(\mathcal{Q}^0_{\text{eval}}, S)$ \Comment{Extract evaluation samples}
\State $\text{acc}_{\text{initial}} \gets \text{Evaluate}(M_{\theta^0}, \mathcal{Q}^0_{\text{sample}})$ \Comment{Evaluate initial model}
\For{$i = 1$ to $T$}
    \State // Step 1: Data generation phase
    \State $\mathcal{C}^{\text{para},i} \gets \{\}$ \Comment{Initialize empty \btt{Augmented-Context} set}
    \For{each $c_k \in \mathcal{C}$}
        \State $\mathcal{C}_k^{\text{para},i} \gets \{\}$ \Comment{Initialize empty set}
        \For{$p = 1$ to $P$}
            \State $c_{k,p}^{\text{para},i} \gets G(\text{``Paraphrase''}, c_k)$ \Comment{Generate paraphrase of $c_k$}
            \State Add $c_{k,p}^{\text{para},i}$ to $\mathcal{C}_k^{\text{para},i}$
        \EndFor
        \State Add all elements of $\mathcal{C}_k^{\text{para},i}$ to $\mathcal{C}^{\text{para},i}$
    \EndFor

    \If{$i = 1$} \Comment{Generate QA only for the first iteration}
        \State $\mathcal{Q}^i \gets \text{GenerateQA}(\mathcal{C})$ \Comment{Generate QA for training}
    \Else
        \State $\mathcal{Q}^i \gets \mathcal{Q}^{i-1}_{\text{eval}}$ \Comment{Use evaluation QA from previous iteration for training}
    \EndIf

    \State $\mathcal{Q}^i_{\text{iter\_sample}} \gets \text{SampleQA}(\mathcal{Q}^i, S)$ \Comment{Samples for intra-iteration evaluation}
\EndFor
\end{algorithmic}
\end{algorithm}

\begin{algorithm}
\caption{PASTA: Auxiliary Functions}
\label{alg:helper_functions}
\begin{algorithmic}[1]
\Function{GenerateQA}{$\mathcal{C}$}
    \State $\mathcal{Q} \gets \{\}$ \Comment{Initialize empty QA set}
    \For{each $c_k \in \mathcal{C}$}
        \For{$j = 1$ to $Q$}
            \State $(q_{k,j}, a_{k,j}^{\text{gold}}) \gets G(\text{``Generate QA pair''}, c_k)$ \Comment{Generate QA pair}
            \State Add $(q_{k,j}, a_{k,j}^{\text{gold}})$ to $\mathcal{Q}$
        \EndFor
    \EndFor
    \State \Return $\mathcal{Q}$
\EndFunction

\Function{SampleQA}{$\mathcal{Q}, S$}
    \If{$|\mathcal{Q}| > S$}
        \State $\mathcal{S} \gets \text{RandomSample}(\mathcal{Q}, S)$ \Comment{Random sampling of specified number}
    \Else
        \State $\mathcal{S} \gets \mathcal{Q}$ \Comment{Use all data}
    \EndIf
    \State \Return $\mathcal{S}$
\EndFunction
\end{algorithmic}
\end{algorithm}

\section{Supplementary Qualitative Evaluation}
\label{appendix:eval_document}

We conducted a qualitative evaluation by examining the actual content during the evaluation of Iteration=5 of the proposed method (Table~\ref{tab:comparison_accuracy}), which achieved the highest Final accuracy. An example is shown in Figure~\ref{appendix:qualitative_results_fujii}.

\begin{figure}[H]
\centering
\begin{minipage}{\linewidth}
    \centering
    \begin{tcolorbox}[
      colback=gray!5,
      colframe=black,
      arc=1.5mm,
      boxrule=0.5pt,
      width=0.98\linewidth,
      left=1mm,
      right=1mm,
      top=0.5mm,
      bottom=0.5mm,
      title={Qualitative Evaluation: Response Evolution Across Learning Stages (Iteration 5)},
      titlerule=0.5pt,
      colbacktitle=gray!20,
      coltitle=black,
      fonttitle=\footnotesize\bfseries
    ]
    \scriptsize
    \colorbox{blue!10}{\tiny\textbf{Context}} \\[-1pt]
    On Billboard JAPAN's "Japan Songs (Country Chart)," Creepy Nuts' "Bling-Bang-Bang-Born" reached \#1 in six countries including the United States, United Kingdom, France, Brazil, South Africa, and Singapore during the April 19-25, 2024 tabulation period. This song is the opening theme for the TV anime "MASHLE: Magic and Muscles - Divine Visionary Candidate Exam Arc." Additionally, Fujii Kaze dominated the top 3 positions in Thailand with "Michiteyuku" at \#1, "Shinunoga E-Wa" at \#2, and "Matsuri" at \#3. Furthermore, six songs by Fujii Kaze ranked in Thailand's top 20.

    \vspace{0.3em}
    \colorbox{blue!10}{\tiny\textbf{Question}} \\[-1pt]
    What are the three song titles that Fujii Kaze used to occupy the top 3 positions in Thailand's chart from April 19 to April 25, 2024?

    \vspace{0.3em}
    \colorbox{green!15}{\tiny\textbf{CPT Response} \textit{[LLM as a judge: Correct]}} \\[-1pt]
    From April 19 to April 25, 2024, Fujii Kaze's songs "Michiteyuku" reached \#1, "Shinunoga E-Wa" reached \#2, and "Matsuri" reached \#3 on Thailand's Billboard JAPAN Japan Songs chart.

    \vspace{0.3em}
    \colorbox{red!15}{\tiny\textbf{SFT Response} \textit{[LLM as a judge: Incorrect]}} \\[-1pt]
    In Thailand's chart from April 19 to April 25, 2024, Fujii Kaze's three songs that occupied the top 3 positions were "Michiteyuku", "Shinunoga E-Wa", and "Matsuri". These three songs ranked at the top positions. \textcolor{red}{Furthermore, "Matsuri" reached \#1. Additionally, "Michiteyuku" reached \#2, and "Shinunoga E-Wa" reached \#3. His songs are highly acclaimed in Thailand's music scene. Moreover, "Matsuri" has been a big hit on Thailand's TikTok. The number of TikTok video posts exceeds 24,000.} [omitted]

    \vspace{0.3em}
    \colorbox{green!15}{\tiny\textbf{DPO Response} \textit{[LLM as a judge: Correct]}} \\[-1pt]
    The three songs by Fujii Kaze that occupied the top 3 positions in Thailand's chart from April 19 to April 25, 2024 were "Michiteyuku", "Shinunoga E-Wa", and "Matsuri".

    \end{tcolorbox}
    \vspace{-6pt}
    \footnotesize\caption{Example of response quality changes across learning stages. The CPT response is accurate, but \textcolor{red}{hallucinations (red text)} appear in the SFT response. After applying DPO, the response improves to a concise and accurate answer based on facts. This confirms that DPO suppresses hallucinations and generates responses containing only necessary and sufficient information.}
    \label{appendix:qualitative_results_fujii}
\end{minipage}
\end{figure}

\section{News Article Categories}
\label{appendix:news_category_list}

\renewcommand{\arraystretch}{1.5}

\begin{table}[H]
    \centering
    \begin{tabular}{|c|c|c|}
        \hline
        \multicolumn{3}{|c|}{\textbf{News Article Categories}} \\
        \hline
        \textbf{Culture \& Entertainment} & \textbf{Economy \& Business} & \textbf{Science \& Technology \& Environment} \\
        \hline
        \textbf{Society \& Incidents} & \textbf{Lifestyle \& Social Trends} & \textbf{Health \& Medical \& Welfare} \\
        \hline
        \textbf{Politics \& International} & \textbf{Sports} & \textbf{Education Issues} \\
        \hline
        \textbf{Disasters \& Accidents} & \textbf{Commentary \& Analysis} & \textbf{-} \\
        \hline
    \end{tabular}
    \caption{Categories of news articles covered in this research}
    \label{tab:news_categories}
\end{table}

Although this research utilized only 128 articles from the \btt{"Culture \& Entertainment"} category, Table~\ref{tab:news_categories} shows the category classification we established when initially scraping web news articles to create our dataset.

\renewcommand{\arraystretch}{1.0}

\section{Experimental Settings Details}
\label{appendix:parameter}
Table~\ref{tab:experimental_settings} shows the details of the experimental settings.
Table~\ref{table:training_settings} shows the settings such as learning rate and batch size used in each learning stage: CPT, SFT, and DPO.

\begin{table}[H]
    \centering
    \caption{Experimental setting details}
    \label{tab:experimental_settings}
    \small
    \begin{tabular}{l|l}
        \toprule
        \textbf{Item} & \textbf{Value} \\
        \midrule
        Target LLM Model & Llama-3.1-8B-Instruct \\
        Target LLM Knowledge Cutoff & December 2023 \\
        Target LLM Max Context Length during training & 1024 Tokens \\
        Target LLM LoRA application layers & All layers including embedding layer \\
        Target LLM Rejected data and evaluation answer generation & Greedy Decoding \\
        \midrule
        GPU Model & NVIDIA H100 \\
        GPU VRAM & 94GB \\
        Number of GPUs & 1 \\
        \midrule
        Library Transformers & Version 4.46.1 \\
        Library TRL & Version 0.12.1 \\
        \midrule
        News article time period & May-July 2024 \\
        News article quantity & 128 articles \\
        \midrule
        Generator LLM for news article dataset processing & gpt-4o-mini-2024-07-18 \\
        Generator LLM for news article dataset processing Temperature & 0.0 \\
        Generator LLM for Knowledge Updating related tasks & gpt-4.1-mini-2025-04-14 \\
        Generator LLM for Knowledge Updating related tasks Temperature & 1.0 \\
        Generator LLM for LLM as a Judge & gpt-4.1-mini-2025-04-14 \\
        Generator LLM for LLM as a Judge Temperature & 0.0 \\
        Generator LLM for Japanese MT-bench++ & gpt-4.1-mini-2025-04-14 \\
        \midrule
        Augmented-Context count per article & 200 \\
        Context-Derived QA count per article & 50 (500 for optimal results in Section~\ref{tab:qa_volume_comparison}) \\
        \bottomrule
    \end{tabular}
\end{table}

\begin{table}[H]
    \centering
    \small
    \caption{Main settings for each learning process}
    \label{table:training_settings}
    \begin{tabular}{lccc}
        \toprule
        \textbf{Hyperparam} & \textbf{CPT} & \textbf{SFT} & \textbf{DPO}\\
        \midrule
        learning rate       & 5e-5  & 2e-5  & 1e-6   \\
        lr scheduler & constant & constant & constant \\
        batch size   & 2      & 8     & 4      \\
        grad accumulation & 64     & 4      & 8      \\
        epoch         & 1      & 1      & 1      \\
        DPO $\beta$               & --     & --     & 0.1    \\
        LoRA Rank $r$            & 128    & 128    & 128    \\
        LoRA $\alpha$               & 128    & 128    & 128    \\
        \bottomrule
    \end{tabular}

\end{table}

\section{Training Steps and Computation Time}
\label{appendix:computation_time}

Table \ref{tab:computation_time} shows the number of training steps and time required for each learning stage. Experiments were conducted on a single NVIDIA H100 GPU.

\begin{table}[h]
    \centering
    \caption{Number of training steps and computation time for each learning stage}
    \label{tab:computation_time}
    \begin{tabular}{lcc}
        \toprule
        \textbf{Learning Stage} & \textbf{Number of Training Steps} & \textbf{Time Required (minutes)} \\
        \midrule
        CPT & 300 & approx. 52 \\
        SFT & 200 & approx. 6 \\
        DPO & 200 & approx. 22 \\
        \bottomrule
    \end{tabular}
\end{table}

The total computation time per iteration was approximately 80 minutes, requiring about 9.3 hours for 7 iterations of training. This time represents only the training time for the \btt{Target LLM} and does not include the time for dataset generation, checkpoint saving, and other processes. The actual execution time needs to account for these additional processes.

\section{Augmented-Context Samples}
\label{appendix:paraphrase_exampl}

Below are sample paraphrases of news articles used in this research.

\begin{tcolorbox}[
  colback=gray!5,
  colframe=black,
  arc=1.5mm,
  boxrule=0.5pt,
  width=\linewidth,
  left=2mm,
  right=2mm,
  top=1mm,
  bottom=1mm,
  title={Original Article},
  titlerule=0.5pt,
  colbacktitle=gray!20,
  coltitle=black,
  fonttitle=\footnotesize\bfseries
]
\small
Music Festival "MUSIC CIRCUS FUKUOKA 2024" to be held overnight at Fukuoka PayPay Dome

The music festival "MUSIC CIRCUS FUKUOKA 2024" will be held on June 8, 2024, at Fukuoka PayPay Dome. This event was first held in Fukuoka in 2018 and is a multi-faceted festival centered around music performances that also offers fashion, food, and sports activities. The 2024 lineup includes Awich, Jin Dogg, OZworld, ALAN SHIRAHAMA (DJ), among others, who will keep the energy high all night. Tickets go on sale from April 1, 2024, priced at 7,000 yen for men and 3,500 yen for women.
\end{tcolorbox}

\begin{tcolorbox}[
  colback=blue!5,
  colframe=blue!30,
  arc=1.5mm,
  boxrule=0.5pt,
  width=\linewidth,
  left=2mm,
  right=2mm,
  top=1mm,
  bottom=1mm,
  title={Paraphrase Example 1},
  titlerule=0.5pt,
  colbacktitle=blue!20,
  coltitle=black,
  fonttitle=\footnotesize\bfseries
]
\small
"MUSIC CIRCUS FUKUOKA 2024" confirmed for all-night event at Fukuoka PayPay Dome

The music festival "MUSIC CIRCUS FUKUOKA 2024" will take place at Fukuoka PayPay Dome on June 8, 2024. First launched in Fukuoka in 2018, this is a hybrid festival featuring music performances alongside fashion, food, and sports activities. The 2024 lineup features Awich, Jin Dogg, OZworld, and ALAN SHIRAHAMA (DJ). Tickets will be available from April 1, 2024, with prices set at 7,000 yen for men and 3,500 yen for women.
\end{tcolorbox}

\begin{tcolorbox}[
  colback=green!5,
  colframe=green!30,
  arc=1.5mm,
  boxrule=0.5pt,
  width=\linewidth,
  left=2mm,
  right=2mm,
  top=1mm,
  bottom=1mm,
  title={Paraphrase Example 2},
  titlerule=0.5pt,
  colbacktitle=green!20,
  coltitle=black,
  fonttitle=\footnotesize\bfseries
]
\small
"MUSIC CIRCUS FUKUOKA 2024" to host all-night event at Fukuoka PayPay Dome on June 8

The combined music, fashion, food, and sports festival "MUSIC CIRCUS FUKUOKA," which debuted in Fukuoka in 2018, will be held at Fukuoka PayPay Dome on June 8, 2024. Performers include Awich, Jin Dogg, OZworld, and ALAN SHIRAHAMA (DJ). Tickets will go on sale April 1, 2024, available for 7,000 yen for men and 3,500 yen for women.
\end{tcolorbox}

These paraphrase examples maintain the factual information of the original article while presenting the same content with different expressions and structures. The paraphrases vary in their organization, simplification of expressions, and stylistic changes, but they accurately preserve important information such as the event date, location, performers, and ticket prices.

\section{Context-Derived QA Samples}
\label{appendix:qa_genelation_exampl}

Below are samples of QA generation based on a news article (the original article from Appendix~\ref{appendix:paraphrase_exampl}).

\begin{figure}[H]
\centering
\begin{minipage}{\linewidth}
    \centering
    \begin{tcolorbox}[
      colback=gray!5,
      colframe=black,
      arc=1.5mm,
      boxrule=0.5pt,
      width=0.98\linewidth,
      left=1mm,
      right=1mm,
      top=0.5mm,
      bottom=0.5mm,
      title={Context-Derived QA Samples: QA about MUSIC CIRCUS FUKUOKA 2024},
      titlerule=0.5pt,
      colbacktitle=gray!20,
      coltitle=black,
      fonttitle=\footnotesize\bfseries
    ]
    \scriptsize
    \colorbox{blue!10}{\tiny\textbf{Question 1}} \\[-1pt]
    How much is the ticket price for men at MUSIC CIRCUS FUKUOKA 2024?

    \vspace{0.3em}
    \colorbox{green!15}{\tiny\textbf{Answer 1}} \\[-1pt]
    The ticket price for men at MUSIC CIRCUS FUKUOKA 2024 is set at 7,000 yen.

    \vspace{0.3em}
    \colorbox{blue!10}{\tiny\textbf{Question 2}} \\[-1pt]
    At which venue in Fukuoka will the MUSIC CIRCUS music festival be held in 2024?

    \vspace{0.3em}
    \colorbox{green!15}{\tiny\textbf{Answer 2}} \\[-1pt]
    The MUSIC CIRCUS FUKUOKA 2024 music event will be held at PayPay Dome in Fukuoka.

    \vspace{0.3em}
    \colorbox{blue!10}{\tiny\textbf{Question 3}} \\[-1pt]
    When do tickets for MUSIC CIRCUS FUKUOKA 2024 go on sale?

    \vspace{0.3em}
    \colorbox{green!15}{\tiny\textbf{Answer 3}} \\[-1pt]
    Tickets for MUSIC CIRCUS FUKUOKA 2024 go on sale from April 1, 2024.

    \vspace{0.3em}
    \colorbox{blue!10}{\tiny\textbf{Question 4}} \\[-1pt]
    Please explain the format of MUSIC CIRCUS FUKUOKA 2024.

    \vspace{0.3em}
    \colorbox{green!15}{\tiny\textbf{Answer 4}} \\[-1pt]
    MUSIC CIRCUS FUKUOKA 2024 is a multi-faceted festival centered around music performances that also includes fashion, food, and sports activities, with events running all night (overnight format).

    \end{tcolorbox}
    \vspace{-6pt}
    \footnotesize\caption{QA samples about MUSIC CIRCUS FUKUOKA 2024.}
    \label{fig:music_circus_fukuoka_qa_samples}
\end{minipage}
\end{figure}

\section{Prompt Content}
\label{sec:appendix_prompts}

Below are the prompts actually used in this research.

\subsection{\btt{Augmented-Context} Generation Prompt}
While the context content and generated content are in Japanese, English prompts were used to enhance instruction adherence.
\label{appendix:Augmented-Context_prompt}

\begin{lstlisting}[language=TeX, basicstyle=\ttfamily\footnotesize, caption={\btt{Augmented-Context} Generation Prompt}, label={lst:augmented-context_prompt}]
## Input Data
<News Article Title>
{title}
</News Article Title>

<Article Content>
{content}
</Article Content>

## Output Format
- JSON Lines (JSONL) ONLY with EXACTLY {n_variations} lines
- Each line MUST contain BOTH a "title" AND a "content" key-value pair
- Each line must be a complete, valid JSON object: { "title": "...", "content": "..." }
- No additional keys, explanations, or blank lines
- Invalid formats (title-only or content-only lines) will be rejected

## IMPORTANT: Format Requirements
- Each line MUST strictly follow this exact structure: { "title": "transformed title here", "content": "transformed content here" }
- NEVER split title and content into separate lines
- NEVER output title-only or content-only JSON objects
- NEVER use trailing commas at the end of JSON objects

## Language Requirement
- The content of all responses (the actual text within the Q and A fields, not the JSONL format itself) must be in Japanese only.

## Output Key Mapping
<News Article Title>  "title"  
<Article Content>     "content"

## Transformation Task
1. Paraphrase each element (news article title, article content) **without changing the meaning at all but with substantial rewording**, generating {n_variations} diverse expressions.  
2. Combine multiple paraphrasing techniques:  
   - Word order changes / active <-> passive voice / formal <-> informal style  
   - Paragraph structure and sentence order rearrangement  
   - Synonym substitution  
   - Variation in punctuation and symbols  
   - Change number/date formats only, not the values themselves  
3. Do not alter proper nouns, numerical values, or dates themselves.  
4. Distribution agencies, reporter names, and other metadata are excluded from transformation.  
5. Markdown and backticks are prohibited.

## Valid Output Example
{ "title":"transformed title here", "content":"transformed content here" }

## Common Errors to Avoid
- DO NOT output title-only lines like: { "title":"transformed title" }
- DO NOT output content-only lines like: { "content":"transformed content" }
- DO NOT split title and content across multiple lines
- DO NOT add any explanation text before or after the JSONL output

## Warning
If you output anything other than **exactly {n_variations} complete JSONL lines** with BOTH title AND content in EACH line, your answer will be considered invalid.
\end{lstlisting}
\subsection{\btt{Context-Derived QA} Generation Prompt}
While the context content and generated content are in Japanese, English prompts were used to enhance instruction adherence.
\label{appendix:prompt_qa_generation}

\begin{lstlisting}[language=TeX, basicstyle=\ttfamily\footnotesize, caption={\btt{Context-Derived QA} Generation Prompt}, label={lst:prompt_qa_generation}]
Please reply in JSONL format only. No other text.

<context>
Article Title: {title}
News article from {date}
{contents}
</context>

<instructions>
1. Read the above context and create {need_qa} diverse question (Q) and answer (A) pairs about the context,
   outputting each pair in JSONL format as {{"Q":"{{question}}", "A":"{{answer}}"}}.

2. Questions (Q)
   - Within 200 Tokens.
   - Ensure that the answer (A) can be uniquely determined from Q alone without the context,
     by including all necessary proper nouns, dates, locations, names, figures, organizations, metrics, etc. specifically and completely.
   - Prohibited: Pronouns or ambiguous expressions like "this article," "last week," or any relative time expressions.
   - Yes/No format questions are prohibited.
   - At least one question must ask about the date of an event, announcement, or occurrence.
     For dates mentioned with only month and day in the article, complete them with the year from context and convert to absolute dates.

3. Answers (A)
   - Within 300 Tokens.
   - Accurately describe the information requested in Q.

4. Language Requirement
   - The content of all responses (the actual text within the Q and A fields, not the JSONL format itself) must be in Japanese only.

5. Diversity and Exclusions
   - Create multiple QA pairs with non-overlapping content.
   - For both Q and A, avoid matching the exact wording of the context sentences; use paraphrasing and structural changes to create diverse expressions.
   - Paraphrasing should maintain the same meaning. Changing the meaning is prohibited (e.g., changing 10 years to 12 years, changing kanji in names, changing proper nouns, etc.)
   - Exclude "article source information" such as publication source, distribution source, media name, reporter name, etc. from proper nouns, and do not include them in QA.

6. Error Handling
   - If the context is corrupted or you cannot create any QA pairs at all, output only one line: {{"Q":"0","A":"0"}}.

7. Output must be JSONL only. Do not include any other text.
</instructions>
\end{lstlisting}
\subsection{News Article Cleaning and Quality Evaluation Prompt}
This prompt is written in English, but a Japanese prompt was actually used.
\label{sec:prompt_cleaning}
\begin{lstlisting}[language=TeX, basicstyle=\ttfamily\footnotesize, caption={Example prompt for news article cleaning and evaluation}, label={lst:prompt_cleaning}]
Please examine the news title (###News Title) and the news article content (###News Article Content) according to the following instructions (###Instructions) and respond strictly in JSON format. The JSON key definitions are listed in ###JSON Definition. Please adhere strictly to these keys.


###Instructions
- Verify if the news title and news article content are consistent. Since these are scraped from the web, they may be inconsistent, or the article content might include unrelated articles or advertisements. This occurs because the structure of websites sometimes includes other news or advertising promotions after or in the middle of the main news article. If you find unrelated news or advertisements mixed in, set the needsCleaning key to True in your response, extract only what appears to be the intended news article, and provide the news title in title and the news content in content. Conversely, if you determine that the current news title and content are appropriate as they are, set the needsCleaning key to False and leave the title and content fields empty.
- Assign a score to the article according to the score definition (###Score Definition) and respond with a single-byte digit of 1, 2, 3, 4, or 5 in the score field. If the needsCleaning key is True, evaluate only the content you extracted as the main news article. If the needsCleaning key is False, evaluate based on the original news title (###News Title) and news article content (###News Article Content).
- After reading the content of this news article, classify it into the most relevant category according to the news article category definition (###News Article Category Definition). The news article category definition is structured as a two-level tree. Enter the first level in maincategory and the second level in subcategory in your response. Be sure to classify it into one of the defined categories. Only if you absolutely cannot classify it into any category, respond with "other" in the relevant part of maincategory or subcategory.


###JSON Definition
- score
- maincategory
- subcategory
- needsCleaning
- title
- content


###Score Definition
- 1: No text. Does not make sense as a text. Random strings, etc.
- 2: Makes sense as text but is not a news article or is completely different text unrelated to the article. Or, does not meet the minimum requirements as an article due to insufficient text length (e.g., the main article is less than 30 characters)
- 3: Borderline quality that barely meets the minimum requirements as a news article. For example, no date so it's unclear when the information is from, no names of subjects or names of objects so it's unclear who or what is being discussed, only the introductory text is available while the important main article is behind a paywall and not scraped, or the available information does not include the basics of news (when, where/who, what happened/what was done) and does not make complete sense as news.
- 4: Meets all the requirements as a news article (when, where/who, what happened/what was done, why) and includes all necessary information.
- 5: In addition to score 4, contains detailed information sufficient for a news article, clearly explaining or illustrating new knowledge or information, some change in the world, new ideas or implications, new discoveries, scientific advances, etc.


###News Article Category Definition
Politics & International:
  - Domestic Politics
  - International Relations
  - Diplomacy
  - Elections
  - Policies

Economy & Business:
  - Macroeconomics
  - Corporate & Industry
  - Finance & Investment
  - Employment & Labor
  - Consumption & Prices
  - Economic Policy
  - New Products & Services
  - Marketing & Advertising

Society & Incidents:
  - Social Issues
  - Crime & Incidents
  - Law & Court
  - Human Rights & Discrimination Issues
  - Education Issues

Lifestyle & Social Trends:
  - Lifestyle
  - Local News
  - Transportation
  - Housing & Real Estate
  - Consumer Issues
  - Trends & Fads
  - Social Conditions
  - Changes in Social Norms
  - Public Opinion Polls

Science & Technology & Environment:
  - Scientific Research
  - Technological Innovation
  - IT & Digital
  - Environmental Issues
  - Climate Change
  - Space Development
  - New Technology Applications

Culture & Entertainment:
  - Arts & Culture
  - Entertainment
  - Sports
  - Food & Cuisine
  - Travel & Tourism
  - Fashion
  - Media

Health & Medical & Welfare:
  - Health & Medical
  - Welfare
  - Aging Society
  - Public Health

Disasters & Accidents:
  - Natural Disasters
  - Accidents
  - Emergencies
  - Disaster Prevention & Mitigation

Commentary & Analysis:
  - Editorial
  - Column
  - Commentary
  - Explanatory Article


###News Title
{news_title}

###News Article Content
{news_content}
\end{lstlisting}

\subsection{News Article Summarization Prompt}
This prompt is written in English, but a Japanese prompt was actually used.
\label{sec:prompt_summary}
\begin{lstlisting}[language=TeX, basicstyle=\ttfamily\footnotesize, caption={Example prompt for news article summarization}, label={lst:prompt_summary}]
Please summarize the following news article (the text between the <news_article> and </news_article> tags) in less than 500 characters.
When summarizing, please observe the following guidelines:
- Do not alter or omit proper nouns (names, place names, facility names, etc.)
- Do not omit dates

<news_article>
{content}
</news_article>
\end{lstlisting}

\subsection{LLM as a judge Prompt}
\label{sec:prompt_llm_judge}
\begin{lstlisting}[language=TeX, basicstyle=\ttfamily\footnotesize, caption={Example prompt for answer correctness evaluation using LLM}, label={lst:prompt_llm_judge}]
<Main Instruction>
Please evaluate the correctness of the response to a question about the context.
The content of eval_answer is the subject of evaluation.
Respond only with a single digit: 0 or 1. Strictly adhere to responding with only 0 or 1. Respond with 0 if incorrect. Respond with 1 if correct.
For detailed evaluation criteria, refer to the "Evaluation Criteria" tag and the "Supplementary Evaluation Criteria" tag.
</Main Instruction>

<Tag Definitions>
Below are the context, question, and response to be evaluated.
The "context" tag section contains the news article context.
The "question" tag section contains the question about the context.
The "eval_answer" tag section contains the response to be evaluated. Evaluate the correctness of this response to the question.
</Tag Definitions>

<Evaluation Criteria>
- The answer is considered correct if it meets ALL of the following conditions. Otherwise, it is incorrect:
- eval_answer is factually correct as a response to the question about the context
- eval_answer does not contradict the content in the context
</Evaluation Criteria>

<Supplementary Evaluation Criteria>
- Dates, times, and quantitative values must match, but the required precision should be determined from the context. For example, Olympic competition times should match to the precision stated in the article (e.g., 0.1 seconds). For events like a change of Prime Minister, matching the day is sufficient.
- Proper nouns such as names of people, places, products, companies, paper titles, etc. must match exactly in content, though expressions may vary. Different formats such as kanji, romaji, hiragana, or English notation are acceptable as long as they refer to exactly the same entity.
- Dates, times, and numerical values can be in different formats as long as they represent the same content. For example, April 1, 2024, is the same as 2024/4/1. 1:00 PM is the same as 13:00 and 1 o'clock in the afternoon. 10,000 yen, 10K yen (using abbreviated thousand notation), and "Ten Thousand Yen" (formal written notation) are the same.
</Supplementary Evaluation Criteria>

<context>
{content}
</context>

<question>
{Q}
</question>

<eval_answer>
{eval_answer}
</eval_answer>

Please strictly adhere to responding with only 0 or 1.
\end{lstlisting}

\end{document}